\documentclass[letterpaper,10 pt,conference]{IEEEtran}
\IEEEoverridecommandlockouts
\usepackage{cite}
\usepackage{amsmath,amssymb,amsfonts}
\usepackage{algorithmic}
\usepackage{graphics} % for pdf, bitmapped graphics files
\usepackage{epsfig} % for postscript graphics files
\usepackage{academicons}

\usepackage{graphicx}
\usepackage[hidelinks]{hyperref}
\usepackage{textcomp}
\usepackage{xcolor}
\usepackage{amsmath}
\usepackage{mathptmx} % assumes a new font selection scheme installed

\usepackage{amssymb}  % assumes amsmath package installed

\usepackage{array}
\usepackage{graphicx}
\usepackage{multirow}
\usepackage{comment}

\newcommand\MyBox[2]{
  \fbox{\lower0.75cm
    \vbox to 1.7cm{\vfil
      \hbox to 1.7cm{\hfil\parbox{1.4cm}{#1\\#2}\hfil}
      \vfil}%
  }%
}

\def\BibTeX{{\rm B\kern-.05em{\sc i\kern-.025em b}\kern-.08em
    T\kern-.1667em\lower.7ex\hbox{E}\kern-.125emX}}
    
\begin{document}

\title{\LARGE \bf Design of Stickbug: a Six-Armed Precision Pollination Robot \\

\thanks{$^{1}$The authors are with the Department of Mechanical and Aerospace Engineering, West Virginia University, Morgantown, WV 26505, USA}
\thanks{$^{2}$Trevor Smith is the corresponding author 
{\tt\small trs0024@mix.wvu.edu}}%

\thanks{This study was supported in part by USDA NIFA Award 2022-67021-36124 "Collaborative Research: NRI: StickBug - an Effective Co-Robot for Precision Pollination," the National Science Foundation Graduate Research Fellowship Award \#2136524, and by the NASA West Virginia Space Grant Consortium, Grant \#80NSSC20M0055.}
}

\author{ $^*$Trevor Smith$^{1,2}$, Madhav Rijal$^{1}$, Christopher Tatsch$^{1}$, R. Michael Butts$^{1}$,\\ Jared Beard $^{1}$, R. Tyler Cook$^{1}$, Andy Chu$^{1}$,  Jason Gross$^{1}$,  Yu Gu$^{1}$ \\
\textit{West Virginia University}\\
Morgantown, USA 
}

\maketitle

\begin{abstract}
This work presents the design of Stickbug, a six-armed, multi-agent, precision pollination robot that combines the accuracy of single-agent systems with swarm parallelization in greenhouses. 
Precision pollination robots have often been proposed to offset the effects of a decreasing population of natural pollinators, but they frequently lack the required parallelization and scalability. 
Stickbug achieves this by allowing each arm and drive base to act as an individual agent, significantly reducing planning complexity.
Stickbug uses a compact holonomic Kiwi drive to navigate narrow greenhouse rows, a tall mast to support multiple manipulators and reach plant heights, a detection model and classifier to identify Bramble flowers, and a felt-tipped end-effector for contact-based pollination.
Initial experimental validation demonstrates that Stickbug can attempt over 1.5 pollinations per minute with a 50\% success rate.
Additionally, a Bramble flower perception dataset was created and is publicly available alongside Stickbug's software and design files.

\end{abstract}

\section{Introduction}

In recent years, evidence of a decreasing population and diversity of wild natural pollinators has caused widespread concern for food security \cite{potts2016assessment}.
These concerns have led to exploring alternative pollination methods to address the growing agricultural demand.
One proposed method to meet these pollination needs is to utilize robotic pollinators.
With these devices, farmers can reduce their dependency on natural pollinators while gaining access to a multitude of potential benefits.
Unlike natural pollinators, which often exhibit specific environmental preferences \cite{https://doi.org/10.1111/gcb.12043}, robotic pollinators can be designed to accommodate different environments and tasks.
This allows robotic pollinators to suit various environmental conditions, including those found in indoor agriculture, such as greenhouses, while being able to accomplish other tasks such as flower thinning \cite{agronomy13112753} and gathering crop data \cite{weiss2011plant,cheein2011optimized,pena2013weed,mueller2017robotanist,gallmann2022flower}.

\begin{figure}
    \centering
    \includegraphics[width=0.48 \textwidth]{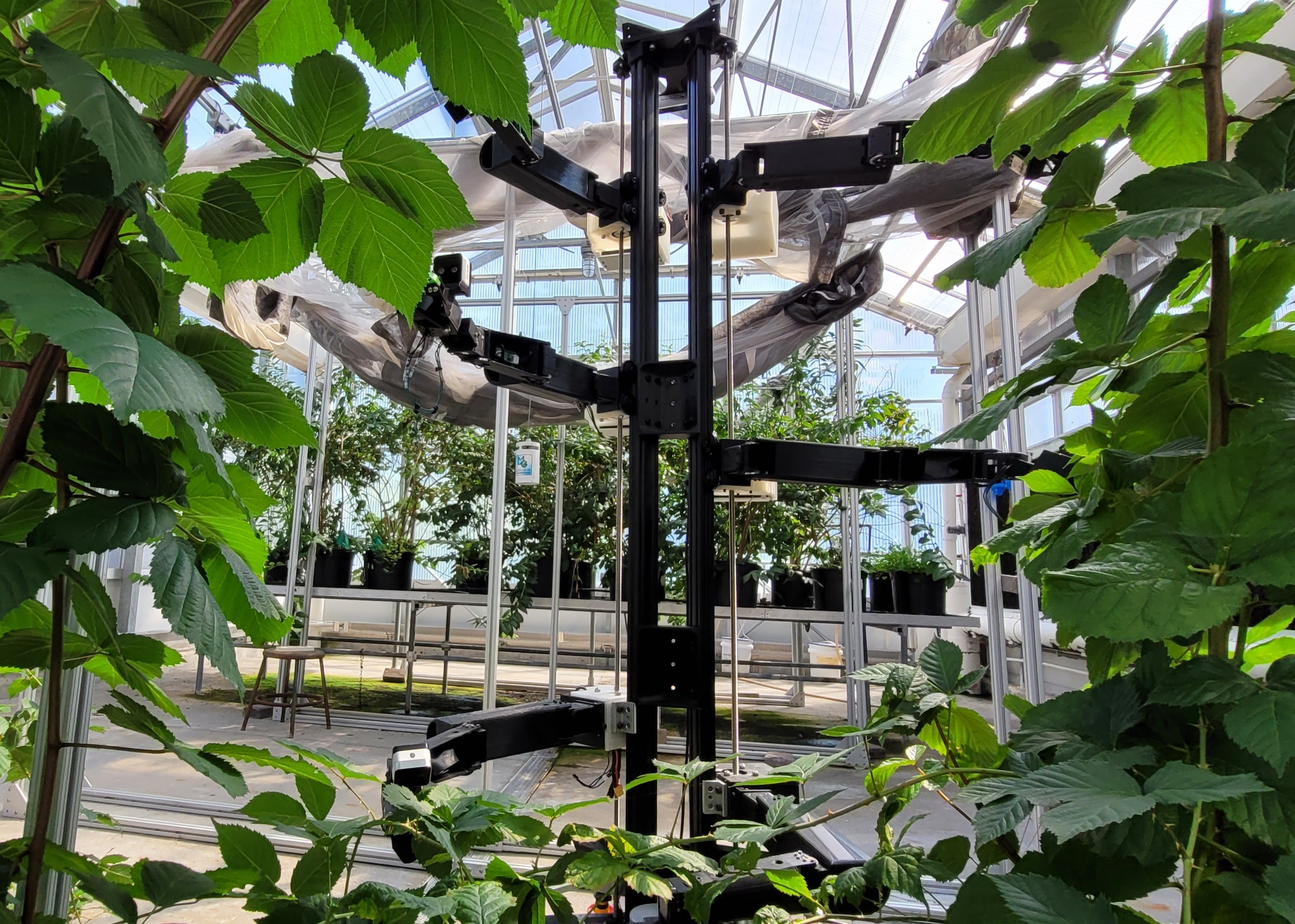}
    \caption{Displays Stickbug, a six-armed precision pollination robot tasked with autonomously navigating, mapping, and pollinating bramble flowers in a greenhouse. Each arm acts as an independent greedy agent that a referee oversees to resolve inter-agent conflicts}
    \label{visual_abstract}
\end{figure}

To accomplish robotic pollination, previous works have utilized non-precise methods that disperse or spray pollen \cite{agronomy13051351}.
These methods have been applied to both ground \cite{KiwiGround} and aerial-based robots \cite{InnovativeandEffectiveSprayMethodforArtificialPollinationofDatePalmUsingDrone,YANG2020101188,9600081,10025186}, but often cannot pollinate specific flower types.
To overcome these limitations, precision pollination techniques have been explored \cite{10417630}, demonstrating pollination in vanilla \cite{doi:10.1504/IJAAC.2013.055096}, kiwifruit \cite{https://doi.org/10.1002/rob.21861}, watermelon \cite{AHMAD2024108753}, tomato \cite{7810939}, apple \cite{ApplePaper}, and tall tree \cite{XIE2023102573} agriculture.
Our previous research has led to the development of a single-armed contact-based precision pollination robot named BrambleBee, designed to autonomously map and pollinate bramble flowers (i.e., blackberry and raspberry flowers) within greenhouse environments \cite{ohi2018design,strader2019flower}.
While BrambleBee demonstrated the utility of mobile precision-based pollination, the system is limited to a single manipulator, hindering the productivity and scalability required to pollinate clusters of bramble flowers.
Other works have explored the potential of small, insect-like flying robots that can be used for pollination \cite{CHECHETKA2017224,Jafferis2019,9786652,hulens2022autonomous}.
While these devices inherit the ability to parallelize the pollination process similar to natural swarms, they often lack the efficiency, maintainability, ease of control, and suite of high-accuracy sensors available on ground-based systems.
An ideal solution would be to design a robot capable of leveraging the advantages offered by both systems.

To achieve this, we introduce the design of Stickbug (shown in Fig. \ref{visual_abstract}), a six-armed pollination robot that bridges the gap between previous precision pollination robots and robotic swarms.
By incorporating multiple robotic arms onto a single drive base, we seek to significantly increase flower pollination throughput.
To overcome the motion planning challenges a large state space presents, each manipulator and the drive-base are designed to operate as independent agents, analogous to a swarm.
However, these agents are overseen by a "referee" who intervenes only to resolve conflicts, ensuring smooth, coordinated operations without constant central control.
Altogether, this work contributes to the literature by:
\begin{enumerate}
    \item Presenting the design of Stickbug, a multi-armed robotic platform tailored for pollinating bramble flowers in greenhouse environments.
    \item  Demonstrating the core hardware and multi-agent autonomy framework with initial pollination experiments.
    \item Sharing the robot design, software, and flower perception dataset. 
\end{enumerate}

The paper progresses as follows: Section \ref{des_req} outlines the design requirements, establishing Stickbug's foundational objectives. Section \ref{des_hard} details the hardware design, focusing on Stickbug's physical components. Section \ref{soft_arch} discusses the robot's software. Section \ref{exp} presents experimental design, results, and analysis. Finally, Section \ref{future} concludes the paper with reflections on the study and potential future directions.

\section{Design Requirements and Challenges}
\label{des_req}

 Individual Bramble plants can grow to more than 2m in height and 50 cm in radius \cite{Huffman2020Brambles, Woodland}, with flower clusters distributed throughout, often obscured by surrounding branches. These plants are arranged in rows inside a greenhouse and collectively produce hundreds of flowers that need individual recognition and tracking. To enable precision pollination, a manipulator must react to the dynamics of the flower's movements, presenting additional challenges to flower tracking and identification. Furthermore, the short blooming period and the need to avoid damaging delicate flowers present time and precision constraints. Therefore, a robotic system with a large workspace, quick and precise manipulators, and the ability to effectively re-identify moving flowers is required.

The robotic system must also be able to navigate, localize, and map the greenhouse environment, which is challenged by tight plant row spacing, GPS signal obstruction, and the lack of a detailed prior map. To address these challenges, the system must be designed with a compact footprint for easier maneuvering and employ alternative localization techniques to GPS for accurate positioning and mapping.

\section{Hardware Design}
\label{des_hard}

\begin{figure}
    \centering
    \includegraphics[width=0.48\textwidth]{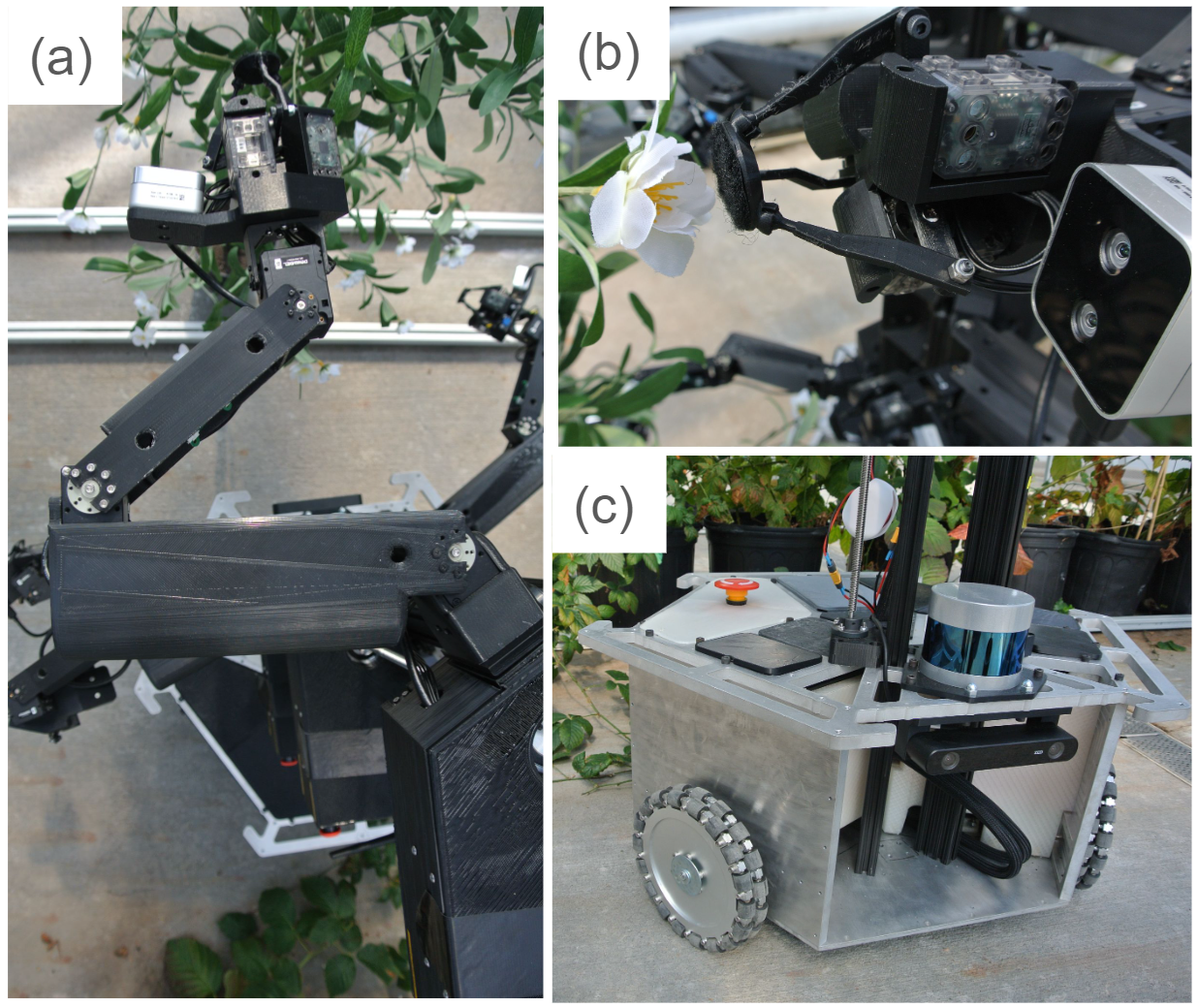}
    \caption{Displays Stickbug's lightweight 3D-printed planar manipulator on the vertical mast left rail (a), pollination end-effector featuring a Realsense D405 depth camera for flower identification and felt tip for pollination (b), and kiwi drive base with a 3D Velodyne LiDAR for navigation and mapping (c).}
    \label{hardware_img}
\end{figure}

 Figure \ref{visual_abstract} depicts Stickbug, a wheeled ground robot featuring six arms, each outfitted with a pollination end effector.  The robot is equipped with multiple sensors, such as a 3D lidar, an Inertial Measurement Unit (IMU), actuator encoders, and a depth camera attached to each manipulator's end-effector. The three main subsystems, drive base, manipulator, and pollination end-effector, are individually displayed in Fig \ref{hardware_img}, and the 3D models are publicly available \cite{Smith_Cook_2024}.

\subsection{Drive Base}

Navigating the narrow aisles of the greenhouse presents significant challenges for robotic mobility. This challenge is addressed by utilizing a Kiwi drive base \cite{killough1991design}, known for its holonomic movement and compactness.  Furthermore, its strafing ability facilitates continuous pollination of bramble plants while navigating through rows. This design makes Stickbug well-suited for spatial constraints and operational demands of greenhouse environments.

 Facilitating navigation and mapping, the drive base, shown in Fig. \ref{hardware_img}, is equipped with a ZED2i for inertial measurements, 3D Velodyne LiDAR, and a Realsense D455 camera. The Realsense camera is placed atop the back of the mast, pointing down to improve immediate obstacle perception and view of areas occluded to the LiDAR. This dense point cloud and inertial data, along with the drive base software, are processed by an Intel NUC Enthusiast equipped with an Nvidia GPU to accelerate computations. Next, to ensure an adequate driving speed, 23cm omni-wheels are driven by 44rpm DC motors with a three-channel Roboteq motor controller, granting a traversal speed of 1m/s. To ensure a low center of gravity, the drive base houses Stickbug's power system consisting of three 12v 42Ahr LiFe batteries. Moreover, it includes an onboard WiFi router for communicating with manipulators.

\subsection{Manipulators}

 Stickbug's six-manipulator design aims to boost pollination efficiency through parallelization but also introduces several design complexities. These complexities include manipulator arrangement, collision mitigation, and cooperation. Moreover, the dense plant row arrangement and tall, bramble plants limit space for multiple large robotic arms. Therefore, Stickbug utilizes a shared vertical mast design to accommodate access to height extremes and six small arms. However, due to the shared mast, collisions become likely, prompting fixed segmented arm workspaces. Yet, the uneven flower distribution demands cooperative manipulator efforts for efficient pollination. To reconcile the need for segmentation and cooperation, Stickbug employs left and right rails on the mast, each supporting three planar manipulators. This arrangement splits the workspace into dynamic lower, middle, and upper regions per rail, simplifying collision avoidance and enabling cross-rail cooperation by creating a shared center workspace. Furthermore, to ensure successful pollination, the manipulator's end effector must be aligned with the flower; thus, a spherical wrist is utilized, omitting the roll joint due to the flowers' radial symmetry. By combining a vertical mast, dual-rail system, small planar arms, and spherical wrist, Stickbug is well-suited to utilize multiple manipulators to pollinate bramble plants in complex greenhouse environments.

To facilitate flower identification and tracking, the manipulators, shown in Fig. \ref{hardware_img} utilize a Realsense D405 depth camera on the end-effector. An onboard Nvidia Jetson then processes the generated color and depth images, manipulator autonomy, and actuator commands. Next, to adequately reach the flowers, Stickbug's mast spans 0.3-1.9m, and each arm has a reach of 50cm from the mast. This creates a 240-degree fanned cylindrical workspace with 120 degrees of overlap centered between the right and left rails. In addition, manipulators employ helix-linear non-captive stepper motors paired with Leadshine closed-loop stepper drivers to enable rail sharing, ensuring precise vertical motion. Furthermore, to minimize dynamic mass, thereby decreasing power needs and enhancing safety, the planar arms and wrist are made from lightweight 3D printed linkages and powered by Dynamixel servos for precise control and integrated wiring.  Moreover, power is provided to the manipulators through daisy-chaining from the drive base.

\subsection{Pollination End-Effector}

Drawing from our previous work \cite{ohi2018design,strader2019flower}, Stickbug addresses the delicate task of flower pollination with a 3 DoF half-Stewart platform end effector, shown in Fig. \ref{hardware_img}, enabling precise pitch, yaw, and forward movements for careful flower brushing. This end effector is constructed from a custom 3D printed design and includes a felt surface for efficient pollen transfer. It is actuated by three Dynamixel XL320 servos, adjusted from the arm's 12V power to 9V via an inline DC-DC voltage regulator, and features integrated color LEDs for autonomy state indication. This configuration ensures Stickbug's gentle interaction with flowers and easy integration with the manipulators.

\section{Robot Software and Autonomy}
\label{soft_arch}

\begin{figure}
    \centering
    \includegraphics[width=0.495\textwidth]{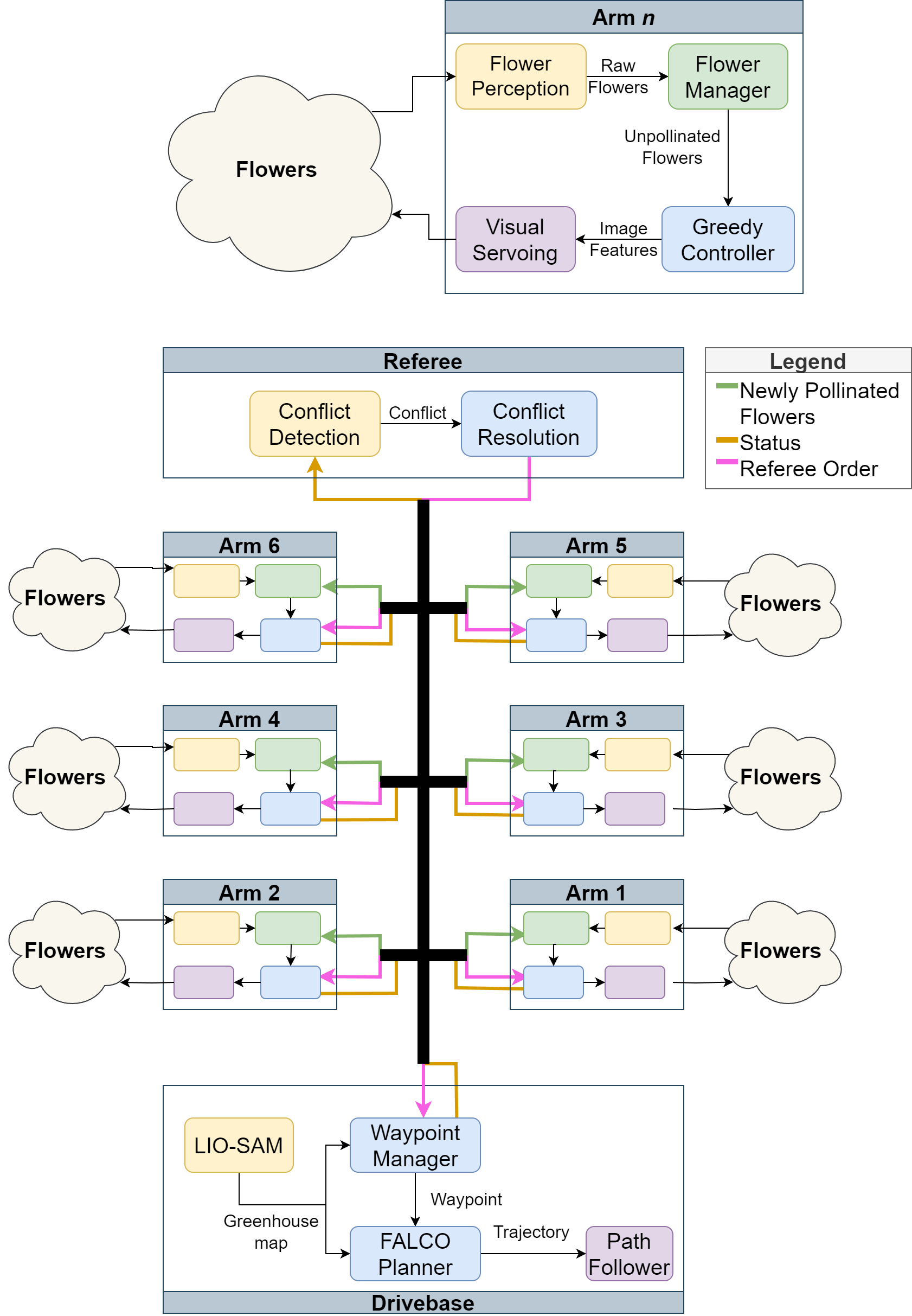}
    \caption{Stickbug's software architecture is distributed across three main agents: the drive base for navigation and mapping, the manipulators for flower identification and pollination, and the referee for inter-agent conflict detection and resolution.}
    \label{soft_arch_img}
\end{figure}

Centrally operating a complex 51 Degree of Freedom (DoF) system presents a formidable challenge due to the computational demands of coordinating multiple manipulators alongside a mobile drive base in the dynamic greenhouse environment. In addition to the practical difficulties of calibrating all manipulators and cameras. To address this, Stickbug's software architecture, displayed in Fig. \ref{soft_arch_img}, adopts a distributed approach, segmenting computation and tasks across the six manipulators and the drive base. This division reduces the system's computational load but introduces the potential for conflicts, such as manipulators competing for the same flowers. Conflicts are addressed by introducing an additional referee agent to maintain some centralized oversight while keeping the computational load manageable. Consequently, Stickbug's architecture is spread across these specialized agents—drive base, manipulator, and referee—with inter-agent communication facilitated by the Robot Operating System (ROS). All software is publicly available \cite{Smith_Tatsch_Rijal_Beard_Butts_Chu_2024}

\subsection{Drive Base}

The drive base agent is tasked with mapping and localizing within the greenhouse where GPS signal is unreliable and navigating the tight aisles to ensure the manipulators can access unpollinated flowers. To overcome GPS obstruction for localization and mapping, LIO-SAM \cite{liosam2020shan}, a Lidar-IMU-based Simultaneous Localization and Mapping (SLAM) algorithm, is utilized. Next, to navigate within the greenhouse, the waypoint manager utilizes a list of waypoints to visit all plants based on a rough prior greenhouse map. The current waypoint is passed to the FALCO path planner\cite{zhang2020falco}, which generates a path using precalculated motion primitives and height-based analysis for local obstacle avoidance. To safely follow the planned trajectory in the narrow greenhouse aisles, the path follower leverages the Kiwi drive's holonomic capabilities and a state machine switching between three driving modes. When the waypoint deviates by over 60 degrees, the robot rotates in place to ensure visibility when traversing due to the mast limiting the LiDAR's Field of View (FoV). For heading differences below 60 degrees with obstacles nearby, translational velocity control ($V_x$ and $V_y$) is used to avoid collisions and maintain a constant heading. In all other scenarios twist control ($V_x$, $V_y$, $\dot{\theta}$) is utilized.

\begin{figure}
    \centering
    \includegraphics[width=0.475\textwidth]{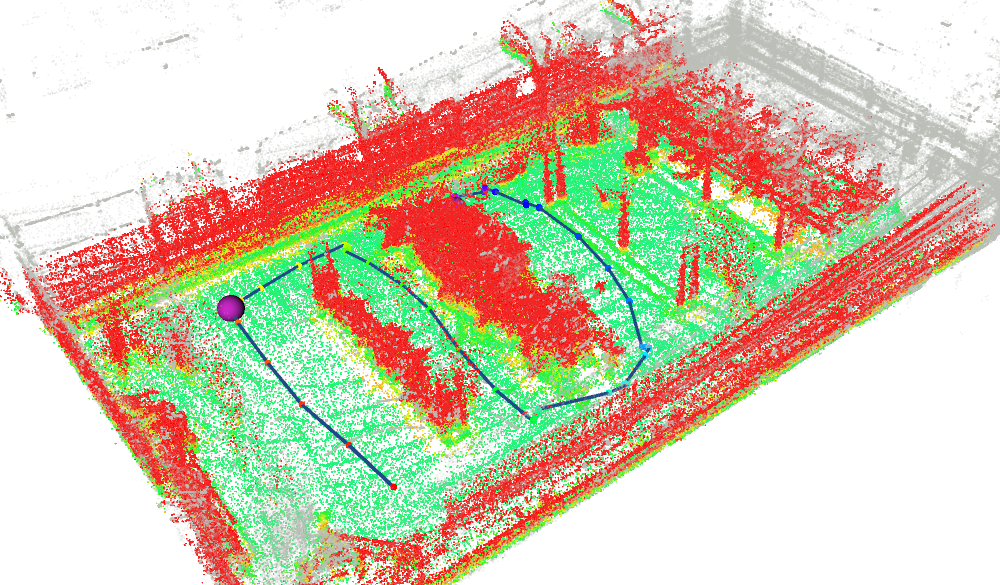}
    \caption{Shows Stickbug navigating greenhouse rows, leveraging LIO-SAM \cite{liosam2020shan} for localization and mapping, and Falco \cite{zhang2020falco} for planning. Green areas are traversable; red indicates untraversable space, and grey areas are outside the scope of local planning. Small spheres mark waypoints, with the current target enlarged in purple. The dark blue line displays Stickbug's path.}
    \label{nav_image}
\end{figure}

The drive base's integration of LIO-SAM, the Waypoint Manager, FALCO, and the Path Follower enables effective navigation and mapping within the greenhouse. Figure \ref{nav_image} shows the robot navigating to predetermined waypoints (marked as small spheres, with the current target enlarged in purple), differentiating traversable (green) from non-traversable areas (red) and regions beyond local analysis (grey). The robot's path, indicated by a dark blue line, demonstrates its adeptness in navigating the constrained and complex greenhouse environment.

\subsection{Manipulators}

Each of the manipulator agents are tasked with identifying and pollinating flowers while coordinating with other manipulators. Together, these tasks present significant challenges as presented in Section \ref{des_req}, such as detecting and locating individual flowers within dense foliage, determining their orientation for adequate pollination, and planning pollination tasks with dynamically moving manipulators whose interactions with plants can increase flower sway, exacerbating the difficulty of identifying and reaching flowers.

\begin{figure}
    \centering
    \includegraphics[width=0.47\textwidth]{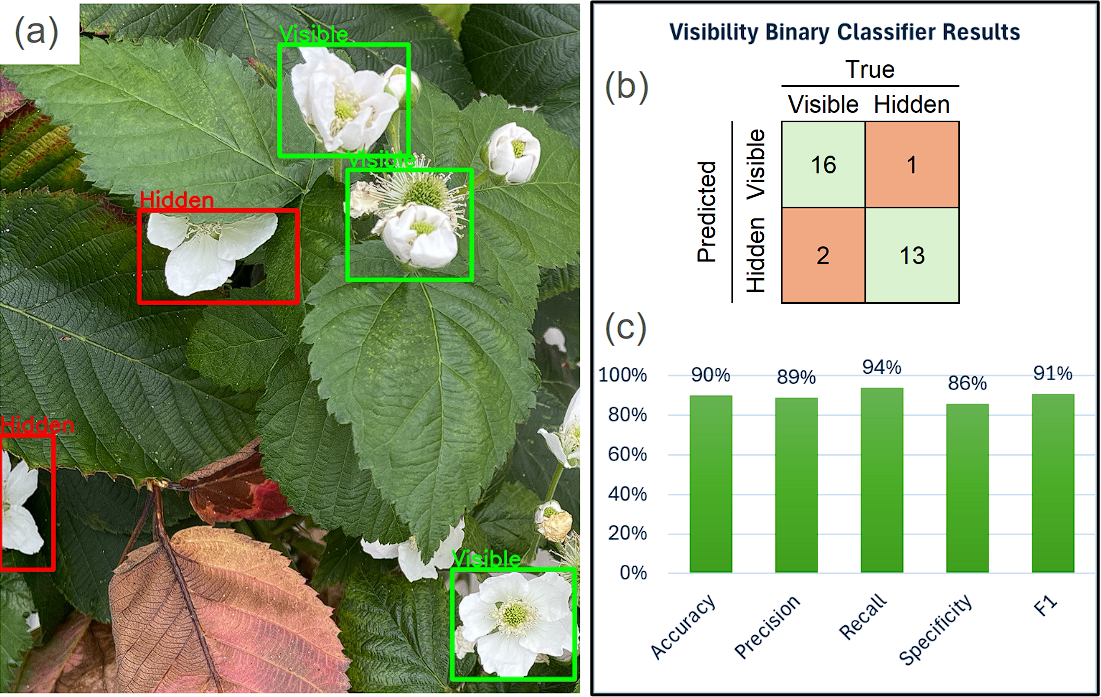}
    \caption{Stickbug employs YOLOv8 \cite{Ultralytics_YOLO_2023} for flower detection, illustrated by bounding boxes, and uses a custom binary classifier to assess the visibility of flower centers for pollination (a). Part (b) presents the confusion matrix of the Flower Center Visibility classifier, with part (c) detailing the classifier's performance metrics.}
    \label{detection_nn}
\end{figure}

 To identify and position individual flowers, the manipulator agents utilize depth cameras and a custom-trained YOLOv8 \cite{Ultralytics_YOLO_2023} model, which extracts bounding boxes from color images to locate each flower's 2D position as shown in Fig. \ref{detection_nn}.a. The depth image is then utilized to estimate their 3D coordinates in the identifying manipulator's base frame. Next, to determine if the camera's current orientation is suitable for pollinating a detected flower, the segmented bounding boxes for each flower are passed to a custom binary classifier that determines if the center of the flower is visible, as shown in Fig. \ref{detection_nn}.a. The flower manager then updates and tracks these flower poses and pollination statuses based on a distance threshold, a method prone to false positives due to clustered flower movements exceeding this threshold. To counteract this, an expiration time is assigned to each flower, extended with each re-identification, effectively filtering out transient detections. This strategy enhances the system's efficiency by focusing on consistently visible flowers while reacting to the dynamic floral environment. 

Next, the manipulator arrangement on the mast grants each arm a unique perspective of the bramble plant, identifying distinct flower sets and establishing an inherent division of labor that enhances coordination. However, this setup also introduces the potential for conflicts in overlapping areas. To mitigate these conflicts, manipulators share their positions and successfully pollinated flower locations, thus avoiding collisions and reducing redundant efforts. However, they refrain from sharing unpollinated flower data, which helps maintain spatial distribution and prevents calibration errors from misleading the coordination efforts.

After identifying flowers, planning the order of pollination in the dynamic environment of moving manipulators and flowers presents a significant challenge. Therefore, a greedy strategy is employed, prioritizing the nearest unpollinated flower, enabling the manipulators to react to environmental changes. Once a flower is selected, visual servoing further aids maneuvering by tracking and compensating for flower movements \cite{chaumette2006visual}. Upon reaching the flower, the end-effector executes gentle brushing motions for adequate pollination, showcasing a reactive approach to overcome the complexities of the dynamic environment.

RGB images of natural and artificial flowers were collected to train the flower perception system with YOLOv8 and the flower center visibility binary classifier. To enhance the diversity of the dataset and simulate real-world conditions, several augmentations were applied: scaling ($\pm25\%$), flipping, cropping (0-30\%), rotating ($\pm20$ degrees), and adjusting saturation ($\pm30\%$), brightness ($\pm30\%$), and noise (0-5\% of pixels). These modifications were implemented using Roboflow software \cite{Robo_Flow} along with all image annotations. Then YOLOv8 and the binary classifier were trained using this dataset, beginning with weights pre-trained on the COCO dataset \cite{lin2014microsoft} for YOLOv8 and random initialization for the binary classifier. The flower data set and trained weights for flower detection and center visibility classification are publicly available \cite{brambledataset}.

After training, the YOLOv8 model achieved a 92\% mean average precision for flower detection. For discerning flower center visibility, the binary classifier demonstrated robust performance on a 32-image test set, as reflected in the confusion matrix (Fig. \ref{detection_nn}.b-c) metrics: accuracy at 90\%, precision at 89\%, recall at 94\%, F1 score at 91\%, and specificity at 86\%. These results underscore the models' adeptness in identifying and localizing the visible centers of bramble flowers, which is crucial for accurate pollination.

\subsection{Referee}

Due to the manipulators' naive decision-making, there will be conflict scenarios in which they interfere with each other's objectives. Thus, the referee agent is tasked with identifying and providing simple resolutions to these conflicts as they arise, thereby maintaining high pollination performance with minimal computational expense. The referee monitors performance and state data from all agents, such as pollination rate, looking for drops in performance that indicate a conflict has occurred and analyzing the agent's state to determine the type of conflict. Once a conflict is identified, the involved manipulator(s) are assigned new poses to resolve it. Currently, the referee focuses on manipulator-manipulator conflicts and manipulator anomalies.

Manipulator-manipulator conflicts occur when two or more arms target positions too close to each other, such as attempting to pollinate the same or adjacent flowers. This leads to a cycle where the manipulators approach, recognize a potential collision, retreat, and then try to approach again. The referee identifies these conflicts by noting when two or more nearby manipulators fail to pollinate within a specific time frame. To resolve this, the referee randomly selects one manipulator to continue while directing others back to their home positions. This approach allows the chosen agent to pollinate the problematic flower, removing it from the unpollinated flower set. The other agent(s) are sent home to redistribute the agents spatially and to broaden their view of the plant, possibly exposing new flowers or encouraging different ones to be pollinated, thus reducing future instances of manipulator-manipulator conflicts.

The referee also addresses manipulator anomalies, such as repetitive unfinished pollination attempts induced by flowers moving in and out of reach or view. These are detected when a manipulator is far from other manipulators and hasn't successfully pollinated within a set time frame. They are then resolved by sending the manipulator back to its home position, similar to manipulator-manipulator conflicts.

\section{Experiments, Results, and Discussion}
\label{exp}

 Experiments were performed to validate Stickbug's design and software, focusing on coordinating multiple manipulators and evaluating Stickbug's current pollination performance.

\begin{figure}
    \centering
    \includegraphics[width=0.28\textwidth]{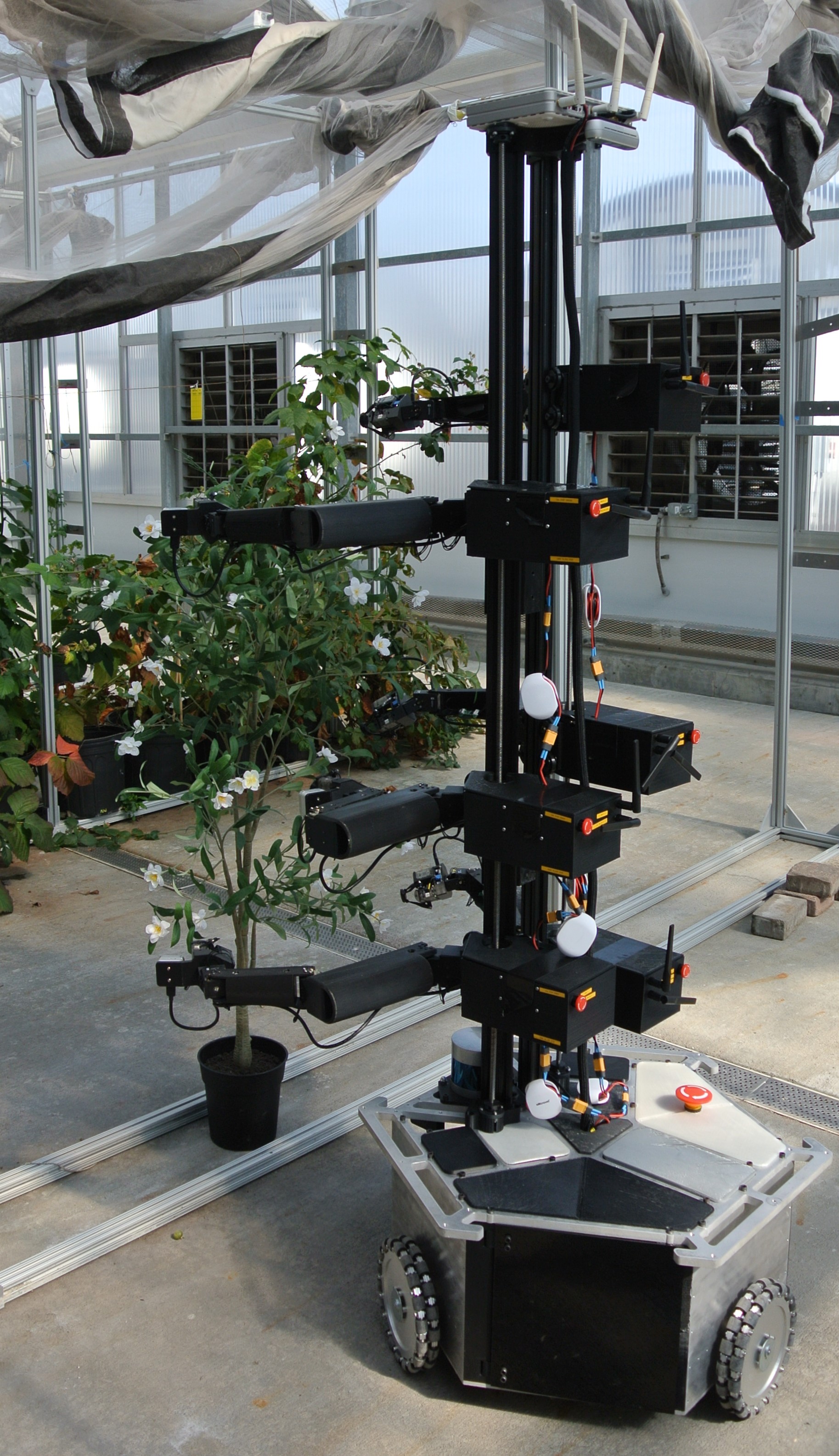}
    \caption{Displays the experimental setup for evaluating Stickbug's design within a greenhouse, featuring an artificial bramble plant with ample flowers for pollination. With the six-armed configuration displayed, this experiment asses manipulator coordination and cooperation across configurations of 1, 2, 4, and 6 arms.}
    \label{exp_set}
\end{figure}

\begin{figure}
    \centering
    \includegraphics[width=0.5\textwidth]{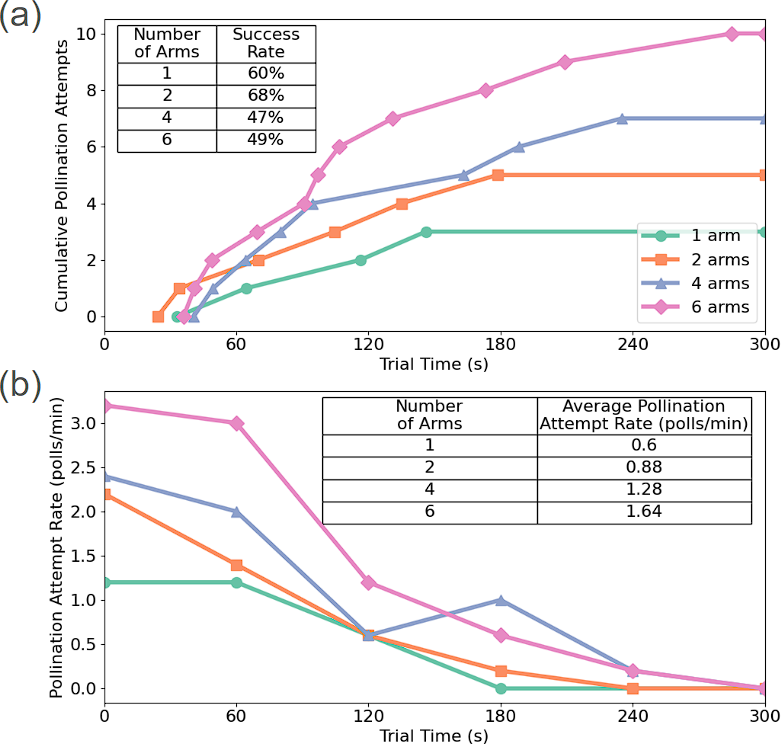}
    \caption{Part (a) shows the cumulative number of pollinated flowers over the experimental window for each set of manipulators. In addition, the pollination rate for each set of manipulators is shown in part (b). As the number of manipulators increased, the total pollination attempts and pollination attempt rate increased.}
    \label{flowersVarms}
\end{figure}

In this experiment, Stickbug was parked in front of an artificial bramble plant, as shown in Fig. \ref{exp_set}, due to the unavailability of natural flowers in this season and was tasked with pollinating as many artificial flowers as possible in five minutes; a human observer recorded the number and time of pollination attempts and successful pollinations. A pollination attempt is defined as the manipulator identifying a flower, moving towards it, and executing the pollination brushing. Given that artificial plants preclude fruit growth as a means to verify pollination \cite{Crassweller2024}, the definition of successful pollination is limited to the end-effector tip making contact with the flower during a pollination attempt. The experiment tested configurations with 1, 2, 4, and 6 manipulators to ensure workspace symmetry beyond the single-arm setup. Each configuration was repeated five times, totaling twenty experiments. The average cumulative number of pollination attempts over the experiment trial and the average percentage of successful pollinations are shown in Fig \ref{flowersVarms}.a. The pollination attempt rate over time and average pollination attempt rate are shown in Fig. \ref{flowersVarms}.b.

Fig. \ref{flowersVarms} reveals that Stickbug, with six manipulators, manages over 1.5 pollination attempts per minute but only achieves a 50\% success rate. This is attributed to manipulators' movements causing flowers to exit the end-effector camera's view, primarily when the camera is close to a flower and its displacement tolerance is limited. Further analysis, as shown in Fig.\ref{flowersVarms}.a, indicates a direct correlation between an increase in the number of manipulators, therefore increasing disturbances, and a decline in success rates. To address this, refining flower tracking techniques and coordinating manipulator actions to minimize disturbances are crucial for improving overall pollination success.

Fig. \ref{flowersVarms} also displays that more arms result in higher rates and total numbers of pollination attempts, indicating efficient spatial coordination despite each manipulator's greedy behavior. This improvement in pollination performance supports the system design of spatially distributing manipulators to inherently allocate tasks and minimize overlap. The lack of manipulator-manipulator conflicts, called by the referee agent, further evidences minimal competition for flowers, reinforcing the effectiveness of this approach in preventing conflicts.

The absence of manipulator-manipulator conflicts among the greedy agents suggests limited cooperation, primarily due to not sharing unpollinated flower locations. This is evidenced in Fig \ref{flowersVarms}, where configurations with four and six arms persist in pollination efforts beyond three minutes—revealing an uneven task distribution with certain manipulators addressing more flowers sequentially rather than all arms pollinating simultaneously. This indicates a design overemphasis on workspace segmentation that hinders manipulator cooperation and highlights the necessity for a more integrated approach to sharing and distributing pollination tasks.

Fig \ref{flowersVarms}.b illustrates a consistent decline in pollination attempts over time, with manipulators ceasing activity after addressing visible flowers. The referee identified this performance drop as a manipulator anomaly and sent them home to reset. However, this strategy, intended to correct failures in pollination attempts, was ineffective in finding new flowers. This underscores the need for search strategies to ensure thorough pollination by detecting and reaching occluded flowers.

\section{Conclusions and Future Work}
\label{future}
This paper introduced the unique design of Stickbug, an autonomous six-armed and multi-agent precision pollination robotic system tasked with pollinating bramble flowers in a greenhouse environment. In addition, we presented the initial demonstration of Stickbug's complex hardware and autonomy stack and achieved over 1.5 pollination attempts per minute with approximately 50\% success rate of pollination on artificial plants in a greenhouse setting. Furthermore, we have created and shared the bramble flower dataset used to train the flower perception, the 3D design files, and the software for Stickbug.

The primary limitation of this study was the lack of access to real flowers for experimentation due to the short pollination season, preventing verification of actual pollination. Secondly, this experimentation revealed the system's limitations in re-identifying and tracking dynamic flower movements, pollination load balancing, and the need for a flower search function to find occluded flowers. Future work will provide subsequent experiments on live plants during the flowering season and improve flower memory and re-identification through Intersection over Union (IoU) and flower relative graph-based mapping. Additionally, we plan to enhance manipulator capabilities by integrating a search function and flower load balancing, leveraging the referee to generate a global flower map to direct manipulators toward unexplored and flower-dense regions.

\section*{Acknowledgment}
We extend our gratitude to Dr. Nicole Waterland and her students for providing access to the greenhouse facility and plants.

\bibliographystyle{IEEEtran}
\bibliography{references}

\end{document}